\documentclass{ieeeaccess}
\usepackage{cite}
\usepackage{amsmath,amssymb,amsfonts}
\usepackage{graphicx}
\usepackage{textcomp}
\usepackage{caption}

\usepackage{algorithm,algcompatible,amsmath}    
\usepackage[noend]{algpseudocode}               
\algnewcommand\INPUT{\item[\textbf{Input:}]}
\algnewcommand\OUTPUT{\item[\textbf{Output:}]}

\def\BibTeX{{\rm B\kern-.05em{\sc i\kern-.025em b}\kern-.08em
    T\kern-.1667em\lower.7ex\hbox{E}\kern-.125emX}}
\begin{document}
\history{}
\doi{}

\title{Location-Aware Box Reasoning for Anchor-Based Single-Shot Object Detection}
\author{\uppercase{Wenchi Ma}\authorrefmark{*},
\uppercase{Kaidong Li}\authorrefmark{*} and \uppercase{Guanghui Wang}\authorrefmark{}, \IEEEmembership{Senior Member, IEEE}}
\address[]{{*}Equal contribution. \\
Department of Electrical Engineering and Computer Science, The University of Kansas, Lawrence, KS, 66045 USA}

\markboth
{Ma \headeretal: Location-Aware Anchor-Based Box Reasoning for Object Detection}
{Ma \headeretal: Location-Aware Anchor-Based Box Reasoning for Object Detection}

\corresp{Corresponding author: Guanghui Wang (e-mail: ghwang@ku.edu).}

\begin{abstract}
In the majority of object detection frameworks, the confidence of instance classification is used as the quality criterion of predicted bounding boxes, like the confidence-based ranking in non-maximum suppression (NMS). However, the quality of bounding boxes, indicating the spatial relations, is not only correlated with the classification scores. Compared with the region proposal network (RPN) based detectors, single-shot object detectors suffer the box quality as there is a lack of pre-selection of box proposals. In this paper, we aim at single-shot object detectors and propose a location-aware anchor-based reasoning (LAAR) for the bounding boxes. LAAR takes both the location and classification confidences into consideration for the quality evaluation of bounding boxes. We introduce a novel network block to learn the relative location between the anchors and the ground truths, denoted as a localization score, which acts as a location reference during the inference stage. The proposed localization score leads to an independent regression branch and calibrates the bounding box quality by scoring the predicted localization score so that the best-qualified bounding boxes can be picked up in NMS. Experiments on MS COCO and PASCAL VOC benchmarks demonstrate that the proposed location-aware framework enhances the performances of current anchor-based single-shot object detection frameworks and yields consistent and robust detection results.
\end{abstract}


\titlepgskip=-15pt

\maketitle

\section{Introduction}
\label{sec:introduction}
\PARstart{D}{eep} networks have been dramatically driving the progress of computer vision, bringing out a series of popular models for different vision tasks \cite{zhang2020self}\cite{xu2020towards}, like image classification \cite{cen2019boosting}\cite{wu2019unsupervised}, object detection \cite{xu2020adaptively}\cite{li2019object}, crowd counting \cite{sajid2020zoomcount}, depth estimation \cite{he2018learning}, and image translation \cite{xu2019adversarially}. Object detection plays an important role and serves as a prerequisite for numerous computer vision applications, such as instance segmentation, face recognition, autonomous driving, and video analysis~\cite{he2017vehicle, ma2018mdcn, bhagavatula2017faster, he2017mask, huang2019mask}. In recent years, the performance of object detectors has been dramatically improved due to the advancement of deep network structure, well-annotated datasets, and effective optimization algorithms~\cite{lin2017focal, zhang2019freeanchor}.

\begin{figure}
    \includegraphics[width=\columnwidth]{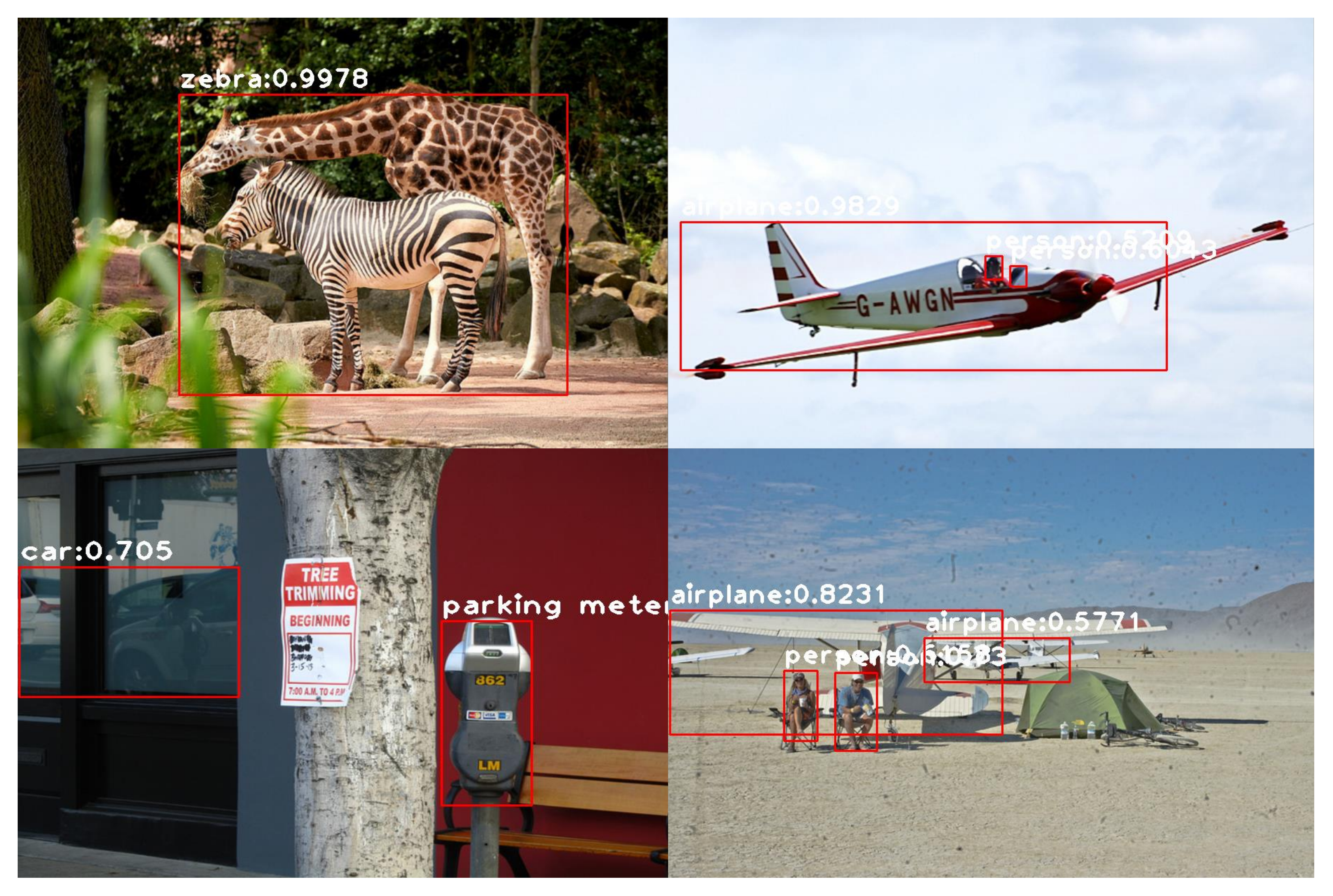}
    \centering
    \caption{Demonstrative detection results on MS COCO~\cite{lin2014microsoft}. The predicted bounding boxes have high classification scores while the localization is misplaced or interceptive. The left two images show misplaced cases in which the zebra is located by a much larger region, and the car is not an actual object. The airplanes in the right two images are partially located, while all of them have high confidence scores. Our method predicts the spatial relation between box proposals and their possible targets so that the interception and misplacement can be minimized.}
    \label{fig:Figure0}
\end{figure}

In this paper, we aim at single-shot object detectors that yield a better trade-off between accuracy and speed, indicating a trend for future frameworks~\cite{liu2016ssd, ma2020mdfn}. We reveal the problem of an inadequate quality criterion for anchor-based bounding box candidates, which is very important for model optimization and detection evaluation. The reason lies in that the quality of bounding boxes should reflect both the spatial location accuracy and the classification probability. While as far as we know, in current deep learning-based object detection pipelines, the scores of the bounding boxes are shared with box-level classification confidence, which is predicted on the proposed features by the classifier. Most importantly, we cannot obtain the location assessment during the inference stage due to the deficiency of labels. It is insufficient to use the classification confidence to measure the bounding box quality since it only serves for distinguishing the semantic categories of proposals, while it is not aware of the assessment towards localization accuracy. The misalignment between classification confidence and bounding box quality is illustrated in Figure~\ref{fig:Figure0}, from which we can see that, although the object instances obtain a high classification confidence score, the box-level localization is not unanimously accurate. If a predicted object is not scored properly, it might be mistaken as a false positive or negative, affecting the NMS process and leading to a decrease of average precision (AP). It is evident that the lack of effective scoring metrics towards the localization quality tends to impair the evaluation.

In this work, we focus on a more reasonable and effective scoring metric for anchor-based bounding box proposals. Different from most previous works that either pursue high-quality classification boxes or focus on score correction working on two-stage object detectors, we demonstrate that there is room of further improvement for popular anchor-based single-shot object detection models by introducing calibrated quality scores that take the location confidence into consideration. Compared with RPN-based frameworks, single-shot object detectors highly depend on qualified box proposals as there is no pre-sift scheme. As anchor-based methods, they are sensitive to location information, which brings challenges for box sifting. To solve this issue, we propose a calibrated quality score (\textbf{CQS}) for each box proposal to realize the location awareness. The localization score indicates the spatial relation to its most-probable target ground truth and ranks the proposals based on the calibrated quality score rather than the classification confidence.

\begin{figure*}[t]
\vspace{-20pt}
\begin{center}
   \includegraphics[width=1.0\linewidth]{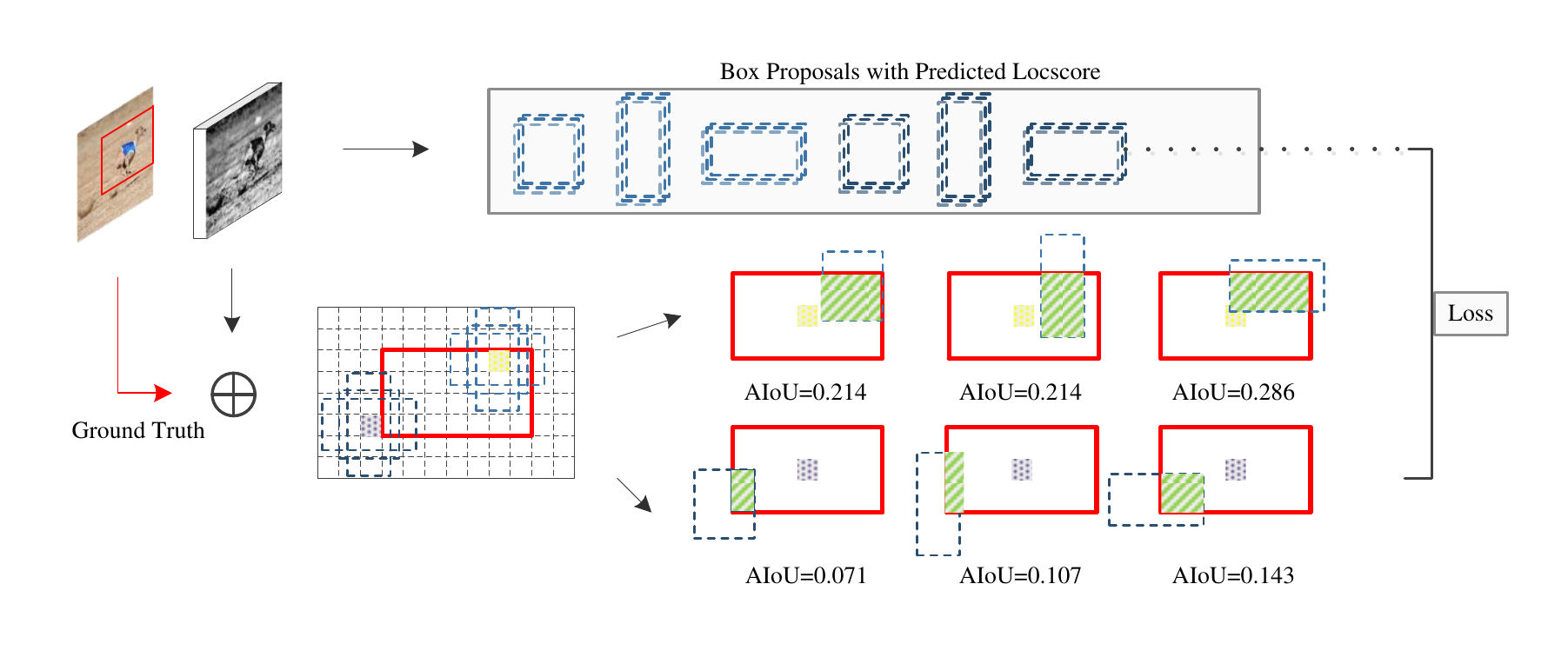}
\end{center}
\vspace{-15pt}
   \caption{AIoU Definition and the Locscore illustration. AIoU as the target of the proposed locscore only needs the input image and its corresponding ground truth. In the fashion of convolution, we evaluate the default boxes with multiple scales at different anchor bases, as the yellow and purple color lumps shown in this figure. For each default box (represented by the blue dotted rectangular), in addition to predicting the shape offsets and the confidence scores as conventional detectors, we also predict the locscore which assesses the possibility of how close the object is to the ground truth. The locscore is learned towards the AIoU calculated by the anchor box and the ground truth, which is denoted by the red angular box in this figure. Specifically, the locscore of a box proposal is learned to match the AIoU between its corresponding anchor box and a certain ground truth box. }
\label{fig:Figure1_1}
\end{figure*}

In anchor-based single-shot object detection, the bounding box proposals are regressed by the space shift relative to the anchors. Thus, the spatial relationship of an anchor and an object ground truth depicts an expectation or estimation of the location relationship between the corresponding box proposal and the target, as depicted in Figure~\ref{fig:Figure1_1}. Inspired by the Average Precision (AP) metric of object detection using pixel-level Intersection-over-Union (IoU) between the predicted bounding box and its ground truth to describe the quality of predictions, we propose a network module to learn the IoU between the anchors and the ground truth directly. For the convenience of discussion, we call it \textbf{AIoU}. We adopt the proposed locscore to learn this AIoU during the training time, and when given the locscore in the test phase, the quality of bounding boxes is reevaluated by integrating locscore into the classification confidence so that the reasoned box proposals are aware of both the location information and the semantic categories. 

Compared with localization and classification regressions that take the ground truths from the labeled dataset, the learning for AIoU only needs to calculate the IoU between the anchors and the ground truths as a target, without further labeling the dataset. Within a detection model, we implement the locscore prediction network as the locscore head, which takes the feature outputs and the calculated AIoU as inputs, and is trained with a common regression loss. We implement object detection experiments with the proposed location-aware anchor-based box reasoning module on popular single-shot object detectors. The results demonstrate that our method can promote the performance of object detection and yield consistent and robust detection results. 
The main contributions of our paper include:
\noindent
	\begin{enumerate}
		\item[C1.] We propose a novel bounding box reasoning method that is aware of the spatial relationship between the box proposals and the probable target ground truth. It is one of the first algorithms that address the issue caused by scoring bounding box proposals only by the classification probability. 
		\item[C2.] This is the first location-aware detection framework designed for the single-shot networks that naturally take the pools of anchor-based box proposals as candidates, ensuring a one-shot learning fashion.  
		\item[C3.] The proposed plug-in locscore head can be integrated with any single-shot detection networks and regressed easily in an end-to-end fashion. By calibrating the detection quality with locscore, the bounding boxes can be penalized if it has high classification confidence while relatively poor localization accuracy. 
		\item[C4.] We demonstrate the effectiveness of the location-aware anchor-based box reasoning scheme through extensive experiments. By introducing the proposed calibrated quality score into the evaluation metric of box proposals, the detection performance is further improved. 
	\end{enumerate}

\section{Related Work}
\subsection{Object Detection}
Multiclass object detection is a core task in the context of deep learning based computer vision projects, which is the joint work of the classification towards contents and the localization towards bounding boxes of instances. Most of these methods adopt the CNN~\cite{lecun1998gradient} based bounding box and classification regressions, followed by a Non-Max Suppression (NMS) algorithm to sift best-qualified box proposals. 

Bounding box regression was first introduced in R-NN~\cite{girshick2014rich}. It enables regions of interest (ROI) to estimate the updated bounding boxes with the purpose of better matching the nearest object instance. Prior works, from Fast R-CNN, Faster R-CNN~\cite{ren2015faster}, R-FCN~\cite{dai2016r}, to YOLO~\cite{redmon2016you}, SSD~\cite{liu2016ssd}, RetinaNet~\cite{lin2017focal}, and RefineDet~\cite{zhang2018single}, have demonstrated that the detection task can be improved with multiple bounding box regression stages~\cite{gidaris2015object}, flexible anchor matching~\cite{zhang2019freeanchor}, the increase of the number of anchors, and the enlargement of the input image resolution, including image pyramids~\cite{lin2017feature}. Among them, the most widely-used and efficient technique is the anchor-based multibox algorithm that can handle scale variation, one of the challenging problems for one-shot object detection. Anchor boxes are designed for discretizing the continuous space of all possible instance boxes into a finite number of boxes with predefined locations, scales, and aspect ratios~\cite{zhu2019feature}. The created instance boxes are regressed to match the ground truth bounding boxes based on the Intersection-over-Union (IoU) overlap, by location shift at certain base anchor with the predefined locations. 

However, there exist underlying limitations. On one hand, the quality of the proposed bounding boxes is only measured by the classification score, leading to the misalignment between the box score and box quality. Due to the unreliability of the box score, a proposal with higher IoU against the ground truth will be ranked with low priority if it obtains lower classification confidence. In this situation, the Average Precision (AP) can be degraded. On the other hand, compared with RPN~\cite{ren2015faster}, the anchor-based technique is more sensitive to box quality especially for the consistency of classification confidence and location accuracy since there is no pre-sift scheme for box proposals in the one-stage case. 

\subsection{Detection Scoring and Correction}
The misalignment of the box score and actual quality has aroused much attention and several correction methods have been proposed in recent years. Tychsen-smith ~\textit{et al.}~\cite{tychsen2018improving} presented a Fitness NMS that corrects the detection score by learning the statistics of best matching detected bounding boxes with the ground truth as a corrective factor. It formulates box IoU statistics prediction as the classification task. It is specifically designed for Denet~\cite{tychsen2017denet}, which restrains its application to arbitrary object detection frameworks. 

Jiang ~\textit{et al}. ~\cite{jiang2018acquisition} proposed a standalone IoU-Net which 
is based on a similar R-CNN structure with a proposal pre-sift scheme to predict IoU between the predicted boxes and the ground truths. It manually designs bounding box filtering as an addition to the data pool of box proposals. The IoU-guided NMS ranks bounding boxes by the predicted localization confidence rather than the conventional classification confidence. Cheng~\textit{et al}. adopted a separate network to correct the scores of samples by processing false-positive samples~\cite{cheng2018revisiting}. SoftMax~\cite{bodla2017soft} proposed to use the overlap between two boxes to correct the low score box. Neumann ~\textit{et al}. ~\cite{neumann2018relaxed} proposed a relaxed softmax to predict the temperature scaling factor in standard softmax for safety-critical pedestrian detection. Both of the two approaches are designed for the two-stage R-FCN based models, relying on the clean proposal data pool. Wu~\textit{et al}. proposed the IoU-aware approach scores the location results while it is merely a RetinaNet based detector~\cite{wu2020iou}, not an arbitrary method.  

Different from the above methods, this study focuses on the essence of the evaluation towards anchor-based box reasoning in single-shot frameworks. We assign each predicted box with a location score by taking aware of the spatial relation between its based anchor and the ground truth. The proposed approach takes both the classification confidence and the location accuracy into consideration to create a complete evaluation of instance box quality so as to narrow the gap between the box score and the actual quality. Furthermore, we build an independent regression branch in the single-shot object detection framework that learns the location confidence specifically and merges this information into the box quality evaluation metric of NMS so as to obtain a more reliable priority ranking. 

\begin{figure*}[t]
\vspace{-20pt}
\begin{center}
   \includegraphics[width=0.9\linewidth]{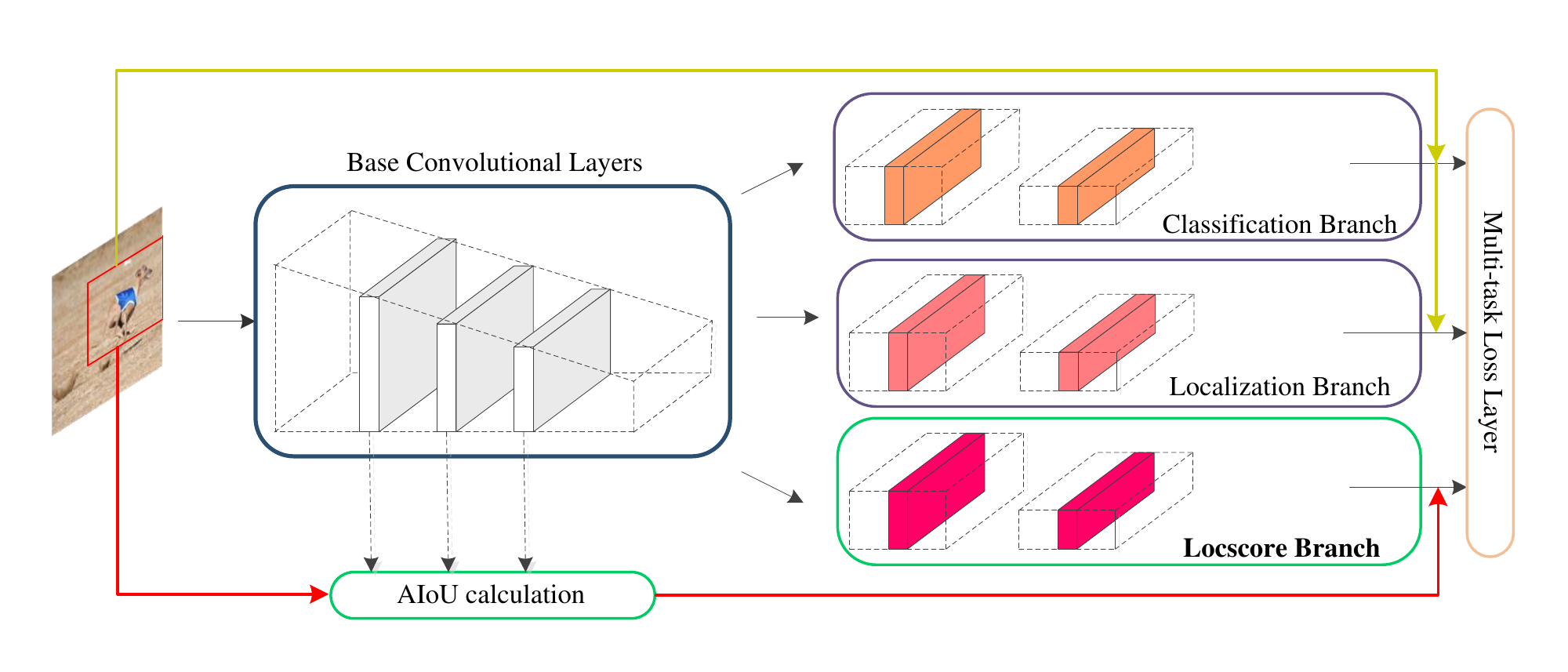}
\end{center}
\vspace{-15pt}
   \caption{The network architecture of object detection with location-aware anchor-based reasoning. The input image is fed into the backbone network to generate feature maps with RoI information. The Locscore Branch is the standard component of the improved model. It takes the output features from the backbone network as inputs and provides a predictive locscore at the end, where its layer structure just follows the one of the classification branch or the localization branch. }
\label{fig:Figure2}
\end{figure*}

\section{Location-Aware Box Reasoning}
\subsection{Motivation}
In current object detection frameworks, the classification and localization regressions are taken as two independent processes. The evaluation towards a detection hypothesis, the detected bounding box, is determined by the highest-ranked element in the classification scores. However, there exist certain situations where the predicted box with a high classification score has low localization accuracy. This kind of hypothesis is harmful in most detection evaluation protocols, such as MS COCO. It is important that a detector can determine when the detection results are trust-worthy and when they are not. This motivates us to integrate the localization score by location-aware anchor-based reasoning for every predicted bounding box based on an anchor position. 

Most previous methods do not consider location confidence as one of the evaluation factors that contribute to the box quality~\cite{tychsen2018improving}\cite{cheng2018revisiting}~\cite{bodla2017soft}\cite{neumann2018relaxed} and the majority of them are designed as specific detectors or merely applied for two-stage detectors with the pre-sift scheme. Although IoUNet obtains competing results with the proposed IoU-guide NMS algorithm that takes localization confidence into consideration, it ranks the boxes only by the localization confidence which highly relies on the clean data pool of proposals produced by two-stage detection models. It is hard to apply it in a single-shot fashion. We propose the location-aware anchor-based box reasoning that focuses on arbitrary single-shot detectors for a better trade-off between accuracy and efficiency.  

We instantiate the location-aware reasoning module by showing how to apply it to the anchor-based single-shot detectors. Without loss of generality, we apply the proposed module to the state-of-the-art RetinaNet and SSD with an additional Locscore head that learns the IoU between the anchor and ground truth, and demonstrates our design from the following aspects: 1) how to realize location awareness for anchors; 2) how to create the branch of location score in the network; and 3) how to generate location-aware anchor-based reasoning during inference time.

\subsection{Location Awareness}
From the perspective of conception, location awareness is simple. In anchor-based detectors, the introduction of location awareness supplements the evaluation towards the quality of the bounding boxes from the perspective of location accuracy. It is realized by learning the IoU between the anchors and ground truths, producing the localization scores. 

\paragraph{Localization Score} We begin with briefly reviewing the evaluation metric towards the bounding box proposals. Following the anchor-based detection, the proposals are created based on anchors with various scales at different positions on the feature maps. The network extracts features from the backbone and performs proposal classification and bounding box regressions respectively. The former one yields confidence scores regardless of the location reference, while the latter one regresses the space migration of candidates. Although the predicted bounding boxes with classification confidence and location prediction, the quality of the box candidates can only be evaluated by the confidence score, without localization assessment. This is due to the lack of ground truth as a reference during the inference stage. Thus, there is a gap between the current metrics and the actual need for evaluating the quality of box proposals. 

We define $P(lc|a_{j})$ as the localization score by learning the pixel-level IoU between bounding box $a_{j}$ and any object ground truth $b_{j}$. We define it as "AIoU". 
\begin{equation} \label{eq:1}
    AIoU = IoU(a_{j}, b_{j})
\end{equation}
In ideal anchor-based detectors, the object is detected by three elements: anchor, box proposal, and object ground truth.  Conventionally, we build the direct correspondence between the anchors and the boxes by regressing space shift, and the relations between the boxes and the object ground truths by matching features. However, there is no direct depiction of the relationship between the anchors and object ground truths, where an underlying location link exists.  The introduction of the location score complements the relational structure of the three essential elements and further calibrates the quality criterion of bounding box proposals by the definition expressed as below. 
\begin{equation} \label{eq:1}
    S(a_{j}) = P(c|a_{j}) \cdot P(lc|a_{j})
\end{equation}
where $P(c|a_{j})$ denotes the confidence score, and $S(a_{j})$ is defined as the calibrated quality criterion of the bounding box proposals. Thus, $S(a_{j})$ should work well on two tasks: indicating the right category that the box belongs to and regressing the IoU of the proposals and the foreground objects. 

\subsection{Localization Score Regression}

\paragraph{Locscore Head} Conventionally, classification and regression are two independent branches for all object categories. Without loss of generality, we introduce the Locscore Head as the third independent branch to predict the IoU between anchors and object ground truths. This head simply follows the layer structure of the other two existing heads so as to save the network characteristics and the advantages of the framework. Thus, it can be implemented as a plug-in module and be integrated with any arbitrary single-shot object detection model. 

The Locscore Head receives the concatenation of features from the output layer of the network as its inputs. It predicts the localization score for each anchor box on the feature maps, which depicts the location relations between the box and the target ground truth. In anchor-based object detection, each predicted bounding box is created based on a certain anchor and regressed location migration. Thus, according to this correspondence among the three elements, each predicted bounding box proposal is corresponding to one anchor so that the box would be assigned a localization score indicating the maximum possibility that it is related to an object from the perspective of localization. 

Based on the analysis above, since the Locscore Head shares the same concatenated features with the other two Heads, classification, and box regression, the predicted three elements have an inner congruent relationship. The Locscore Head, thus, can be taken as an independent regression branch and treated as an individual learning task.

We define the Locscore Loss to regress the Localization Score. It follows the loss definition for classification regression. Then the Locscore Head is integrated into an anchor-based object detection framework, and the whole network can be trained end-to-end. Specifically, we define the classification loss and box regression loss as $L_{cl}$ and $L_{bb}$, respectively. In addition, we introduce locscore loss $L_{lc}$ as another penalty item to the cost function, as shown below, 
\begin{equation} \label{eq:2}
    L^{*} = \lambda_{1} * L_{cl} + \lambda_{2} * L_{bb} + \lambda_{3} * L_{lc}
\end{equation}
where $L^{*}$ denotes the final loss function. In all experiments, we adopt equal weights for the three loss items in consideration for stability, so that $\lambda_{1}$ = $\lambda_{2}$ = $\lambda_{3}$ = 1. The locscore loss forms a lower bound in the space localization, and by training, we further pull down this lower bound.

\subsection{Box Reasoning during Inference}

\paragraph{Calibrated Quality Score (CQS)} We define and propose a calibrated quality score by introducing the localization confidence into the assessment of the predicted bounding box proposals. Thus, the quality score is disintegrated into two data spaces, where both classification confidence and location accuracy are taken into consideration, as shown in Equation \ref{eq:1}. This CQS during inference time becomes a new criterion for the sifting of qualified candidates. It complements the defect of independent classification and location regressions that lead to the deficiency of localization reference during inference time. 

\paragraph{Inference} At the inference stage, the proposed candidates with coordinates, conferences, and localization scores are integrated into the non-maximum suppression (NMS) algorithm. Different from its classical counterparts, the NMS in our model does not rank the box proposals merely by classification probabilities in the first step. Instead, we use the CQS as the ranking criterion so as to push the box candidates to indicate the spatial relations with potential objects, in terms of the initial idea that the quality of bounding boxes is tightly correlated to both the spatial information and classification confidence. We assume the saved boxes after the above sifting are qualified candidates. But in order to weaken the sensitivity towards the less qualified proposals for the single-shot models, we adopt the confidence cluster~\cite{jiang2018acquisition} to further enhance the reliability of the sifted boxes by updating the confidence score $S_{i}$ of box $i$ with $S_{i}=max(S_{i}, S_{j})$, where $j$ indicates box $j$ that is deleted by box $i$ in NMS. Details are shown in Algorithm 1. 

Specifically, suppose the network outputs $N$ bounding boxes, the NMS firstly rank them by the proposed CQS , then we follow the same procedure to remove the candidate boxes which overlap each other over a threshold of $\epsilon=0.5$. At last, the top-k scored boxes are selected and fed into the output head to generate multi-class boxes. 

\begin{algorithm}
\caption{Location-Aware Box Reasoning. Classification confidence and localization score are independently regressed during the training time but the two values are taken as combined consideration during the inference time when evaluating the anchor-based box proposals. }
\begin{algorithmic}[1]
    \INPUT $B$, $P_{c}$, $P_{lc}$, $\epsilon$. 
    \item[$B$: set of anchor-based bounding box proposals.]
    \item[$P_{c}$: classification confidence by mapping $f_{c}$.]
    \item[$P_{lc}$: localization score by mapping $f_{lc}$.]
    \item[$\epsilon$: IoU threshold in NMS.]
    \OUTPUT $D$, set of detected boxes with classification confidence $P_{c}$. 
    \State $D \leftarrow \varnothing$ 
    \While{$B$ $\neq$ $\varnothing$}  
        \State $S$ $\leftarrow$ $P_{c}$ $\times$ $P_{lc}$
        \State $b_{m}$ $\leftarrow$ $\arg \max_{i} S(b_{i})$
        \State $S_{m} \leftarrow S(b_{i})$
        \State $P_{m}$ $\leftarrow$ $P_{c}(b_{i})$ 
        \State $B$ $\leftarrow$ $B \setminus b_{m}$
        \For{$b_{j} \in B$}
        \If {IoU($b_{m}$, $b_{j}$) > $\epsilon$}
        \State $B$ $\leftarrow$ $B \setminus b_{j}$
        \If {classification cluster}
        \State $P_{m}$ $\leftarrow$ $max(P_{c}(b_{j}), P_{m})$.
        \EndIf
        \EndIf
        \EndFor
        \State $D \leftarrow D \cup \{\left \langle b_{m}, P_{m} \right \rangle\}$
    \EndWhile  
\end{algorithmic}
\end{algorithm}

\section{Experiments}
We conduct experiments on the detection tasks of the MS COCO~\cite{lin2014microsoft} and PASCAL VOC~\cite{everingham2010pascal} datasets. MS COCO contains 80 object categories, we follow COCO 2017 settings, using the 115k images \textit{train} split for training, 5k \textit{validation} split for results analysis. The COCO results are reported by its evaluation metrics AP (Average Precision over IoU thresholds), including, AP@0.5 (IoU equals to 0.5), AP@0.75 (IoU equals to 0.75), AP (averaged on AP over IoU thresholds from 0.5-0.95 with a step size of 0.05), AP$_{S}$, AP$_{M}$, AP$_{L}$ (AP at different scales of objects).  In the VOC experiments, we follow the same practice as in the literature, the models are trained on the union of PASCAL VOC2007 and 2012 trainval set (16,551 images) and tested on PASCAL VOC 2007 test set (4952 images). The overlap threshold for each one of the 20 categories in VOC is set to 0.5. 

\subsection{Implementation Details}
We adopt the consistent location-aware box reasoning (LAAR) framework for all experiments. We use the ResNet-50 based FPN network and VGG-16 based SSD network for ablation study, respectively. For ResNet-50 FPN, the input images are resized with a minimum 608px along the short axis and a maximum of 1024px along the long axis for both training and test. We train the network and choose the model at the epoch that yields the best performance. The learning rate is reduced by the Plateau strategy, the same as that of the original RetinaNet. For SSD, the input images are resized to 300$\times$300 for both training and test, as the common rule in literature. The rest of all configurations are identical to the realization in~\cite{liu2016ssd}. We train the network for 120,000 iterations and decrease the learning rate after 80,000 and 100,000 iterations. The optimizer for RetinaNet experiments is Adam with an initial learning rate of 0.00001, and for SSD experiments is SGD with momentum 0.9. In the test, all the results are evaluated by the NMS, where the top-100 score detection is retained for each image. 

\paragraph{Learning Scenarios} In order to identify the gains of locscore regression constraint and location-aware box reasoning respectively, we intentionally design independent learning scenarios where we do solo locscore regression constraint without quality score calibration, the complete LAAR detection framework with CQS, and the complete LAAR detection with CQS and confidence cluster. We list them as follows:

\begin{itemize}
 
    \item {\em Independent Locscore Constraint ($ILC$):} We introduce the locscore regression during training time while do not consider the predicted locscore to calibrate the classification score during the inference period.
    \item {\em Locscore Constraint with CQS ($LC$):} We conduct complete location-aware box reasoning with the given detector, which means we add locscore regression constraint during training and introduce the calibrated quality score by the predicted locscore during the test time. 
    \item {\em Locscore Constraint with CQS and classification cluster ($LC+CS$):} Based on LC, we introduce the classification cluster after calibrating quality score by the predicted locscore, as shown in Algorithm 1.
\end{itemize}

\subsection{Ablation Studies}
We evaluate the contribution of one important element to our location-aware box reasoning for object detection, the constraint brought by the Localization Score Regression.

\paragraph{Locscore constraint for better optimization} Compared with conventional object detection frameworks, we introduce the additional constraint term in the loss function by doing the localization score regression. As far as we know, this is the first time to directly explore the relationship between the anchors and the ground truths in single-shot fashions. The proposed procedure explores the predefined prior information of anchor boxes and the ground truths from the perspective of spatial location. As is known, anchor-based fashion defines fixed anchor positions and their multiple-ratio variations on a feature map, which means for a certain image, there exist predefined location relations between the anchor boxes and the ground truths. We introduce this relation in the penalty function to help constrain the classification and localization regressions. Especially for the latter one during training, when the coordinates are learned in a deflected direction, there exists a correction by the locscore regression. This constraint results in better optimization as demonstrated in Table \ref{tab:1} and \ref{tab:2}. 

In Table \ref{tab:1}, the ILC version leads the original RetinaNet in most cases. For AP$_{0.75}$, ILC improves the detection accuracy by a rough 1\% than the original ResNet.  Although the ILC falls behind AP$_{S}$, it leads by large margins in both AP$_{M}$ and AP$_{L}$. In Table \ref{tab:2}, we obtain consistent results. In this VOC setting, we list the AP results for all the 20 categories. The ILC shows better mAP than the original SSD and yields clearly higher AP for most categories. These results support the idea that the introduction of locscore regression yields effectively positive constraint towards the other two regressions. Thus, the locscore branch boosts the mutual promotion of classification and localization, helping to improve the situation when confidence score and localization accuracy are opposite, such as when confidence score is 0.6 and localization accuracy is 0.2 compared to that when confidence score is 0.2 and localization accuracy is 0.6, which guarantees the feasibility of the proposed approach.\\


\begin{table}[t]
	\begin{minipage}[b]{\linewidth}
	    \centering\footnotesize
        \begin{tabular}{c|cccc}
        \hline
                                & R-Net  & R-Net+ILC &  R-Net+LC & R-Net+LC+CS\\ \hline
            Backbone & R-50 & R-50 & R-50 & R-50 \\ 
            AP        & 30.4 \%  & 30.9 \% & 30.7 \%  & \textbf{30.9} \%  \\ 
            AP$_{0.5}$   & 47.3\%  & \textbf{47.7} \% & 47.2\%  & 47.4 \%  \\ 
            AP$_{0.75}$   & 32.1 \%    & 33.0 \% & 32.9\% & \textbf{33.2} \%  \\ 
            AP$_{S}$   & \textbf{13.9}\%  & 13.0 \% & 12.9 \% & 13.0 \%  \\ 
            AP$_{M}$        & 33.1 \%    & \textbf{34.0} \% & 33.8\% & 33.9 \%  \\ 
            AP$_{L}$       & 43.7\%    & 44.1 \% & 44.2\% & \textbf{44.3} \%  \\ \hline
        \end{tabular}
         \vspace{3mm}
        \caption{\footnotesize The mAP of RetinaNet on COCO val2017. R-50 indicates ResNet50 with FPN and R-Net refers to RetinaNet.}
        \label{tab:1}
	\end{minipage}
	\vspace{-10pt}
\end{table}

\begin{table}[t] \label{tab:2}
	\begin{minipage}[b]{\linewidth}
	    \centering\footnotesize
        \begin{tabular}{c|cccc}
        \hline
                                & SSD & SSD+ILC & SSD+LC & SSD+LC+CS\\ \hline
            Backbone & VGG-16 & VGG-16 & VGG-16 & VGG-16 \\ 
            aeroplane        & 82.97 \%  & 81.87 \% & 82.43 \% & 82.41 \% \\ 
            bicycle   & 84.18\%  & 83.61 \% & 83.13\% & 83.18 \% \\ 
            bird   & 75.34 \%    & 75.23 \% & 74.23\% & 75.56 \% \\ 
            boat   & 70.98\%  & 70.35 \% & 70.62 \% & 70.71 \% \\ 
            bottle        & 50.44 \%    & 51.97 \% & 51.62\% & 51.90 \%  \\ 
            bus       & 84.29\%    & 86.05 \% & 84.49\% & 86.03 \%  \\ 
            car       & 86.32\%    & 85.30 \% & 84.60\% & 85.53 \%  \\ 
            cat       & 88.12\%    & 88.36 \% & 87.45\% & 88.26 \%  \\ 
            chair     & 61.54\%    & 62.18 \% & 58.82\% & 62.02 \% \\ 
            cow       & 79.94\%    & 83.03 \% & 82.28\% & 83.18 \% \\ 
            diningtable    & 77.12\%    & 75.80 \% & 72.40\% & 76.20 \%  \\ 
            dog    & 85.05\%    & 84.31 \% & 82.38\% & 84.13 \% \\ 
            horse   & 87.60\%    & 87.03 \% & 85.92\% & 87.68 \%   \\ 
            motorbike    & 82.84\%    & 82.95 \% & 81.83\% & 83.50 \%   \\ 
            person     & 78.98\%    & 79.09 \% & 77.71\% & 79.12 \%   \\ 
            pottedplant    & 52.17\%    & 51.94 \% & 50.27\% & 51.45 \%  \\ 
            sheep    & 78.61\%    & 77.40 \% & 75.40\% & 77.05 \%  \\ 
            sofa    & 78.33\%    & 80.29 \% & 77.49\% & 80.12\%  \\ 
            train   & 87.88\%    & 86.65 \% & 84.80\% & 87.71\%  \\ 
            tvmonitor    & 76.33\%    & 77.29 \% & 74.93\% & 77.10\%   \\ 
            mAP     & 77.45\%    & 77.53 \% & 76.14\%  & \bf{77.64}\%  \\ \hline
            
        \end{tabular}
         \vspace{3mm}
        \caption{\footnotesize mAP of SSD on VOC2017. The category name indicates its corresponding AP result.}
        \label{tab:2}
	\end{minipage}
	\vspace{-20pt}
\end{table}

\begin{figure*}[t]
\begin{center}
   \includegraphics[width=1.0\linewidth]{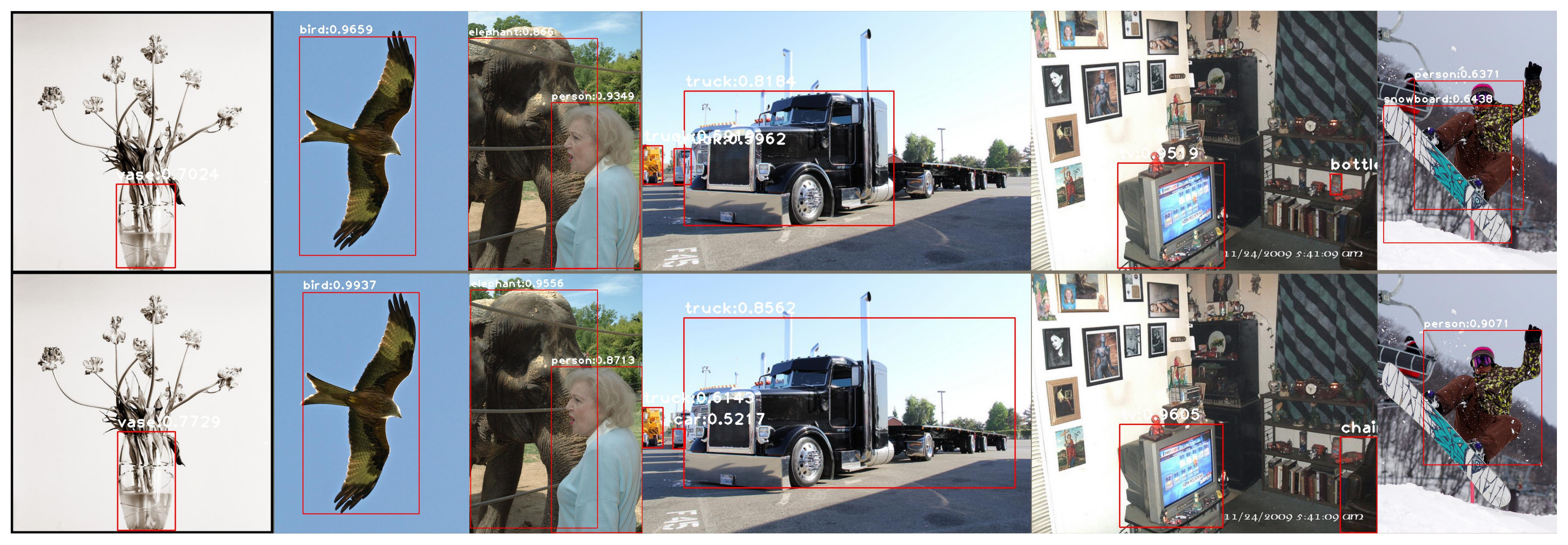}
\end{center}
\vspace{-15pt}
\caption{COCO cases compared between RetinaNet (1st row) and RetineNet+LC (2nd row), where LC version achieves higher localization accuracy.}
\label{fig:Figure3_betterbox}
\end{figure*}

\begin{figure*}[t]
\begin{center}
   \includegraphics[width=1.0\linewidth]{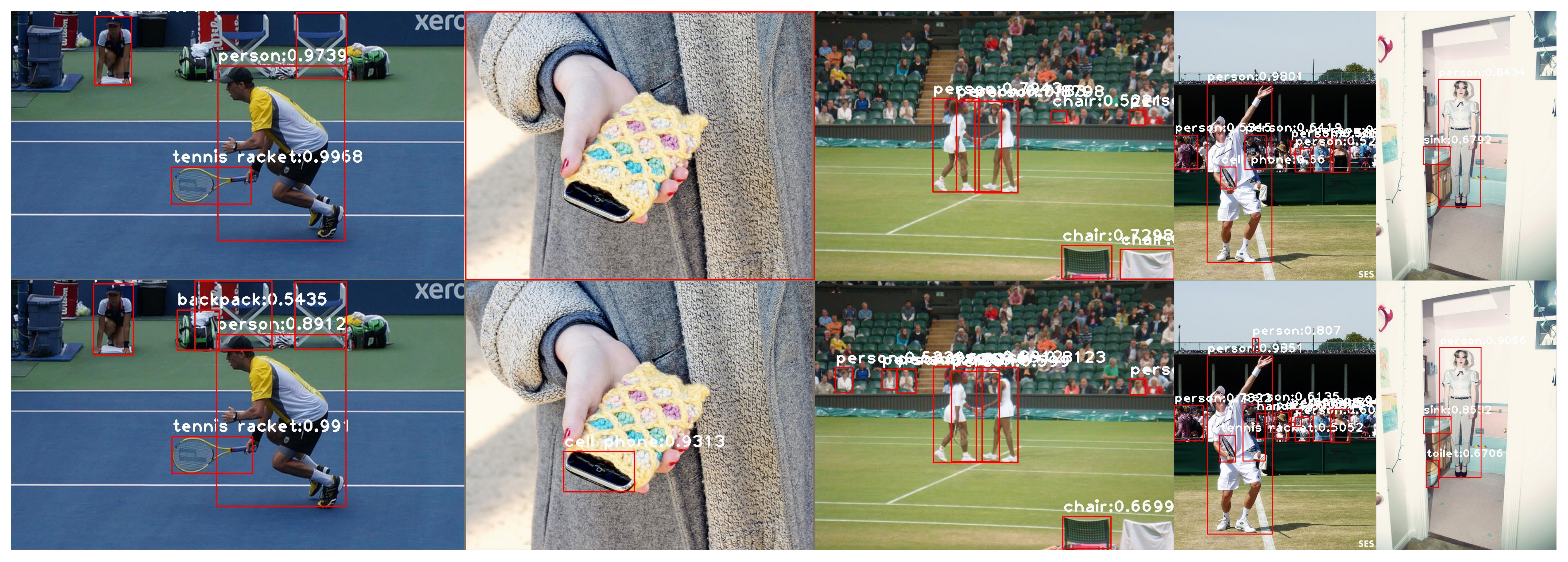}
\end{center}
\vspace{-15pt}
\caption{COCO cases compared between RetinaNet (1st row) and RetineNet+LC (2nd row), where LC version performs better for hard objects.}
\label{fig:Figure3_hard}
\end{figure*}

\begin{figure*}[t]
\begin{center}
   \includegraphics[width=1.0\linewidth]{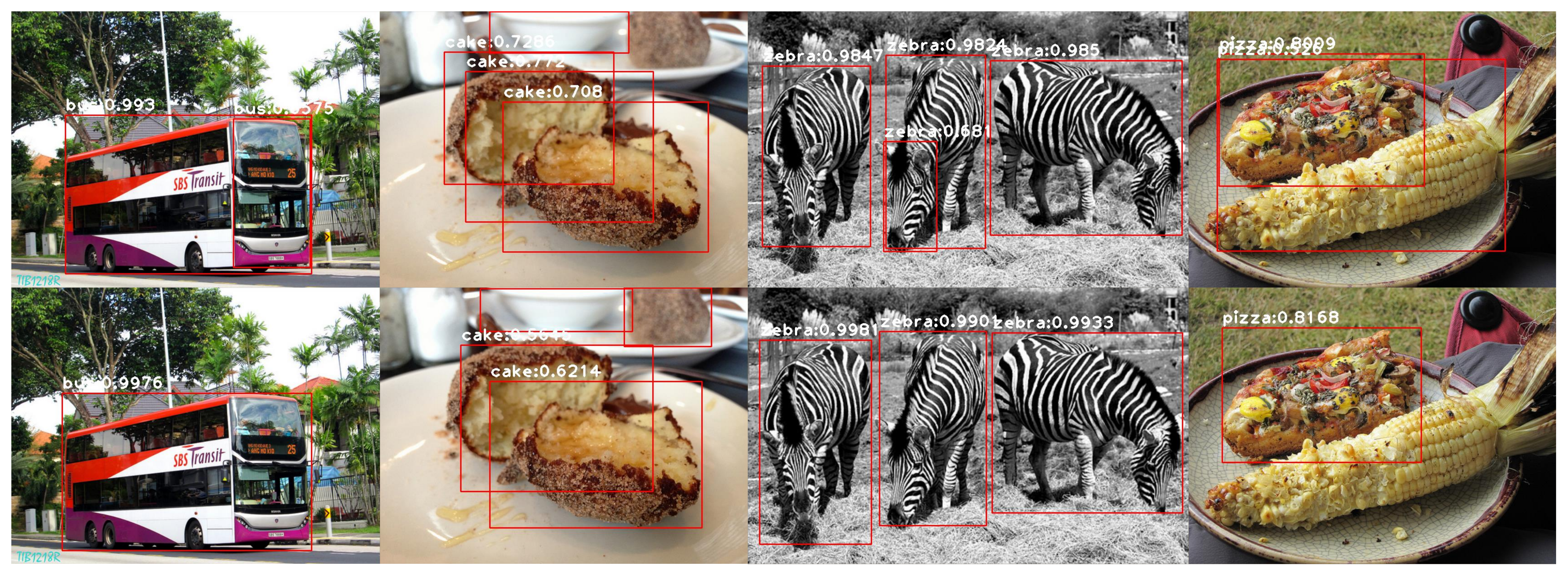}
\end{center}
\vspace{-15pt}
\caption{COCO cases compared between RetinaNet (1st row) and RetinaNet+LC (2nd row), where LC version has better filtering for low-quality boxes.}
\label{fig:Figure3_multibox}
\end{figure*}

\subsection{Quantitative Results}
We extensively evaluated the proposed method with two popular detectors, SSD and RetinaNet, on Pascal VOC 2007 and COCO val2017, respectively. We also compared the proposed algorithm with state-of-the-art methods, SoftMax~\cite{tychsen2018improving} and IoUNet~\cite{jiang2018acquisition}. Since these methods are confined to the application of R-CNN based two-stage frameworks, they are highly dependent on the pre-sift scheme by RPN, where the proposal pool could already be regarded as clean data. The proposed method aims at anchor-based single-shot detection models without the pre-sift scheme, where the anchor-based produced boxes can be regarded as rough candidates. We conduct experiments and verify that we can not directly introduce the above algorithms in the single-shot fashion as a comparison, like the one proposed by IoUNet by ranking the boxes using localization confidence. It leads to a sharp drop in mAP. It is understandable that the produced boxes by anchors are rough candidates that are unreliable for quality ranking. Therefore, these methods can not be directly applied to single-shot models. 

To make a fair comparison, we integrate the core idea of IoU-guided NMS produced by IoUNet to merge the classification cluster into our algorithm and form the `LC+CS', described in lines 11 and 12 of Algorithm 1. We then conduct comparison and show quantitative results in Table \ref{tab:1} and \ref{tab:2}. From Table \ref{tab:1} we can see that, compared with RetinaNet, LC  model achieves stable improvement in most cases. Specifically, LC obtains better AP, AP$_{0.75}$, AP$_{M}$, and AP$_{L}$. For AP$_{0.75}$, LC achieves an enhancement of 0.8\%, and for AP$_{M}$, LC improves by a margin of 0.7\%. We can conclude that, for the detection accuracy with high request and objects with the most common sizes, our method exhibits clear advantages. In Table \ref{tab:2}, the LC version does not lead the ranking although its corresponding ILC version performs better than the original SSD model. This reflects the fact that the anchor-based single-shot models are sensitive to location accuracy and the produced rough candidates have less reliable location outputs than the expectation. From Tables \ref{tab:1} and \ref{tab:2} we can see that `LC+CS' achieves the best results in both experiments, with over 1\% enhancement in some cases, such as in RetinaNet at AP$_{0.75}$. The results demonstrate that the introduction of classification clustering could compensate for the uncertainty caused by the location outputs. 

\subsection{Comparison and Discussion}
In this section, we first discuss the quality of the predicted bounding boxes and investigate the upper bound of the performance of the LAAR model, and analyze the benefits of locscore learning. Here, all the results are evaluated on COCO2017 validation set using RetinaNet and ReinaNet with the LC models. 
\subsubsection{Fitter and tighter bounding boxes}
In Figure \ref{fig:Figure3_betterbox}, It is evident that the LC model predicts higher-qualified boxes than RetinaNet, and most of the boxes are tighter. Specifically, the boxes for the vase, the eagle, and the truck, are more accurate, and the box for the chair shows better performance in occluded cases. This demonstrates that the introduction of locscore learning can improve the accuracy of bounding boxes, and help select the best proposals that have the maximum alignment between the quality score and box quality. Tighter boxes in practice can help clear up some current dilemmas in industry. 
\subsubsection{Better for hard objects}
In Figure \ref{fig:Figure3_hard}, it can be seen that the LC model is able to detect harder objects, like the small backpacks, the occluded cellphone, and the persons in the audience. These small or occluded objects are easy to be overlooked during the suppression process, as their classification scores can be small due to the deficiency of the effective features. While with the locscore, the detection score can be calibrated as these hard objects could probably have high localization scores if they are labeled. Thus, the actual detection accuracy can be raised by calibrating the box quality score especially when the original classification confidence is low. 
\subsubsection{Waiving low-quality boxes}
In addition, from Figure \ref{fig:Figure3_multibox}, we can see the LC model is better at sifting out some low-qualified boxes, like the smaller bus box, the redundant cake, and zebra box. Similarly, this can be explained by the calibrated quality score. Although some boxes' classification scores are high, if their location scores are low, these boxes can still be regarded as low-quality objects, which could be disregarded in NMS.

\section{Conclusion}

This paper reveals the problem of object detection score as one of the primary limitations of current anchor-based single-shot object detectors. To address this issue, we have proposed the localization score (locscore) regression and location-aware box reasoning, where the classification score is aligned with the predicted locsocre so that the localization accuracy is taken into the assessment of the quality of the bounding box proposals, which has been overlooked in most popular object detection frameworks. Extensive experimental results show that the proposed approach can consistently improve the detector's performance to yield reliable bounding boxes. The proposed module can be directly applied to any single-shot object detection models to improve their performance in both classification and localization.

\section*{Acknowledgement}
The work was supported in part by USDA NIFA under the grant no. 2019-67021-28996, NASA EPSCoR under the grant no. 80NSSC20M0160, and the Nvidia GPU grant.

{\small
	\bibliographystyle{IEEEtranS}
	\bibliography{access}
}


\EOD

\end{document}